\documentclass[letterpaper, 10 pt, journal, twoside]{IEEEtran}
\usepackage{graphicx}
\usepackage{amsmath} 
\usepackage{threeparttable}
\usepackage{array}
\usepackage{float}
\usepackage{subfigure}
\usepackage{cite}
\usepackage[numbers,sort&compress]{natbib}
\usepackage{array} 
\usepackage{multirow}
\usepackage{amsfonts}
\usepackage{booktabs} 
\usepackage{amsthm} 

\usepackage{hyperref} 
\hypersetup{
	colorlinks=true,
	linkcolor=blue,
	filecolor=blue,      
	urlcolor=blue,
	citecolor=blue,
}
\usepackage{bm}
\usepackage{color}
\usepackage[table]{xcolor}

\ifCLASSINFOpdf
\else
\fi
\begin{document}
%
\title{VisionSafeEnhanced VPC: Cautious Predictive Control with Visibility Constraints under Uncertainty for Autonomous Robotic Surgery}
%
%
%


\author{Jiayin Wang$^{1,3}$, Yanran Wei$^{*2}$, Lei Jiang$^{3}$, Xiaoyu Guo$^{4}$,~\IEEEmembership{Member,~IEEE}, Ayong Zheng$^{3}$, Weidong Zhao$^{1}$, Zhongkui Li$^{2}$,~\IEEEmembership{Senior Member,~IEEE}%
\thanks{This work was supported by the National Natural Science Foundation of China under grants 62425301, U2241214, 62373008, and T2121002.}
\thanks{$^{1}$School of Computer Science and Technology, Tongji University, 200092, Shanghai, China.}%
\thanks{$^{2}$School of Advanced Manufacturing and Robotics, Peking University, 100871, Beijing, China.}%
\thanks{$^{3}$MicroPort MedBot (Group) Company Ltd., 201203, Shanghai, China.}%
\thanks{$^{4}$Department of Biomedical Engineering, City University of Hong Kong, 999077, Kowloon, Hong Kong.}%
\thanks{\textsuperscript{*} Corresponding author.}
\thanks{E-mail:\{jaryerwang, wd\}@tongji.edu.cn, \{yanranwei, zhongkli\}@pku.edu.cn, \{leijiang, ayzheng\}@microport.com, xiaoyguo@cityu.edu.hk.}}

%
%

\markboth{IEEE Robotics and Automation Letters.}
{WANG \MakeLowercase{\textit{et al.}}: Cautious Visibility-Constrained Control for Autonomous Robot-Assisted Surgery} 

%



\maketitle
\begin{abstract}
Autonomous control of the laparoscope in robot-assisted Minimally Invasive Surgery (MIS) has received considerable research interest due to its potential to improve surgical safety. Despite progress in pixel-level Image-Based Visual Servoing (IBVS) control, the requirement of continuous visibility and the existence of complex disturbances, such as parameterization error, measurement noise, and uncertainties of payloads, could degrade the surgeon's visual experience and compromise procedural safety. To address these limitations, this paper proposes VisionSafeEnhanced Visual Predictive Control (VPC), a robust and uncertainty-adaptive framework for autonomous laparoscope control that guarantees Field of View (FoV) safety under uncertainty. Firstly, Gaussian Process Regression (GPR) is utilized to perform hybrid (deterministic + stochastic) quantification of operational uncertainties including residual model uncertainties, stochastic uncertainties, and external disturbances. Based on uncertainty quantification, a novel safety aware trajectory optimization framework with probabilistic guarantees is proposed, where a uncertainty-adaptive safety Control Barrier Function (CBF) condition is given based on uncertainty propagation, and chance constraints are simultaneously formulated based on probabilistic approximation. This uncertainty aware formulation enables adaptive control effort allocation, minimizing unnecessary camera motion while maintaining robustness. The proposed method is validated through comparative simulations and experiments on a commercial surgical robot platform (MicroPort MedBot$^\circledR$ Toumai$^\circledR$) performing a sequential multi-target lymph node dissection. Compared with baseline methods, the framework maintains near-perfect target visibility ($>99.9\%$), reduces tracking errors by over $77\%$ under uncertainty, and lowers control effort by more than an order of magnitude.
\end{abstract}

%
\IEEEpeerreviewmaketitle

\section{Introduction}
\IEEEPARstart{R}{obot-assisted} Minimally Invasive Surgery (MIS) has emerged as a major trend in modern surgical practice due to its ability to enhance safety, reduce patient trauma, and improve procedural convenience~\cite{saeidi2022autonomous}. Within this paradigm, the laparoscope serves as the surgeon’s “eyes”, and its coordination with the instrument-holding arms is essential for maintaining a clear and stable Field of View (FoV) throughout the procedure. In current clinical workflows, laparoscope adjustments are predominantly performed manually~\cite{zhang2024automatic}, forcing the surgeon to alternate between controlling the laparoscope and the instruments~\cite{wang2023robotic}. The manual paradigm disrupts operational continuity, impedes real-time coordination, and increases the risk of losing the FoV during critical phases of surgery~\cite{peng2022endoscope}. These challenges are particularly pronounced in complex multi-target surgical tasks—such as sequential lymphatic dissections—where safe, autonomous, and coordinated laparoscope control becomes crucial for surgical success~\cite{zhang2023visual}.

Visual servoing has been widely adopted as a foundational method for autonomous laparoscope FoV adjustment~\cite{de2022autonomous,ribeiro2023second}. Image-Based Visual Servoing (IBVS) employs image features to perform target-tracking motion control, offering rapid response and computational simplicity, making it suitable for scenarios reliant on 2-Dimensional (2D) image data~\cite{9126170}. Position-Based Visual Servoing (PBVS) employs 3-Dimensional (3D) pose estimation to deliver enhanced spatial positioning accuracy~\cite{yu2023position,huang2022review}. While both approaches can maintain continuous target visibility, their inherent reliance on pixel-level target tracking causes the laparoscope to passively follow instrument motions, lacking proactive, task-driven, and surgically coordinated FoV trajectory planning. This is especially problematic for sequential multi-target procedures with spatially dispersed targets, where such passivity reduces operational precision and task efficiency~\cite{lin2022robust,sun2021visual}. Moreover, in practical surgical robotic systems, complex disturbances—such as parameterization errors, payload uncertainties, unmodeled residual dynamics, stochastic measurement noise, and even actuator faults—can degrade visual stability and compromise procedural safety~\cite{10144490}.
\begin{figure*}[tp]
	\centering 
\includegraphics[width=1\linewidth]{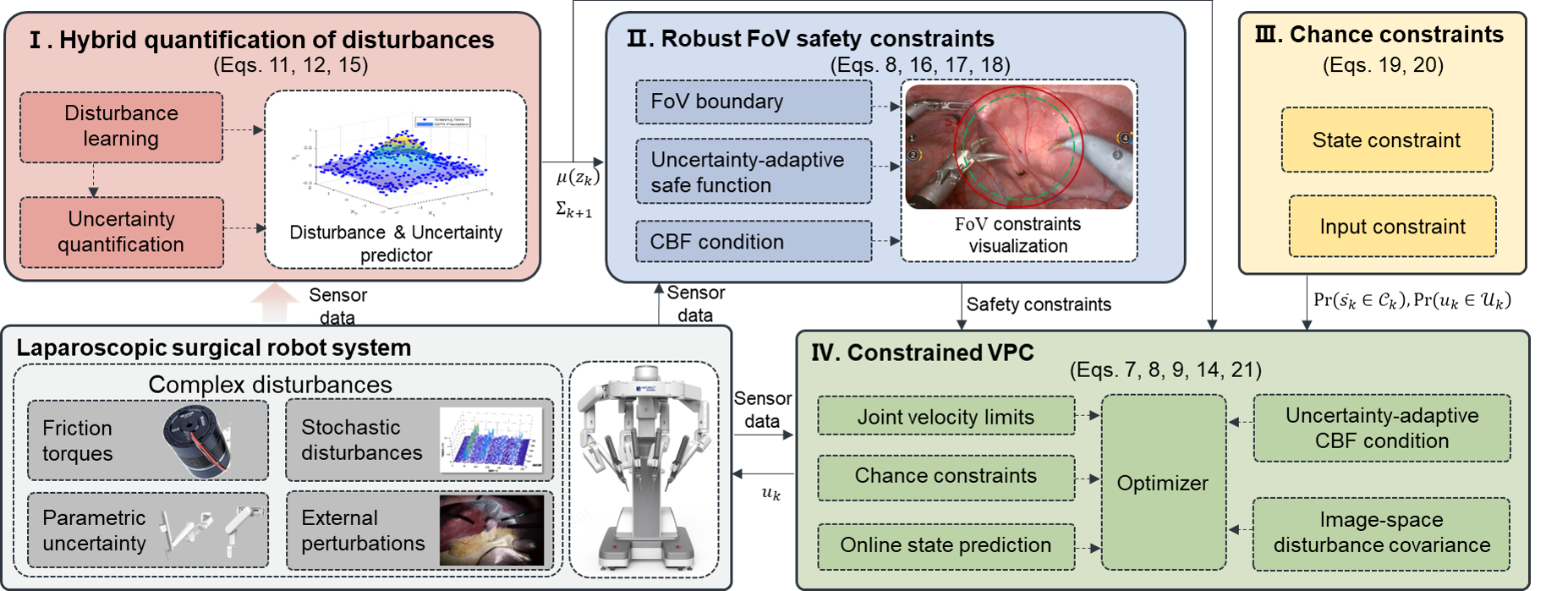}
\caption{The framework of the proposed VisionSafeEnhanced VPC. (I) \emph{Hybrid quantification of disturbances}: disturbance learning via GPR provides mean compensation $\mu(z_k)$ and predictive covariance $\Sigma_{k+1}$ for uncertainty propagation. 
(II) \emph{Robust FoV safety constraints}: the CBF-based formulation incorporates uncertainty-adaptive margins to guarantee continuous instrument visibility. 
(III) \emph{Chance constraints}: probabilistic bounds are imposed on both state and input using Gaussian approximations, ensuring constraint satisfaction under uncertainty. 
(IV) \emph{Constrained VPC}: integrates joint velocity limits, chance constraints, and uncertainty-adaptive CBFs into an online optimizer, achieving safe and robust trajectory planning. 
The laparoscopic robot system generates sensor data subject to complex disturbances (e.g., parametric uncertainty, friction, stochastic noise, and external perturbations), which are processed by the framework to ensure safe and efficient laparoscope motion.}
\label{framework}
\end{figure*}

To address these limitations, online motion planning within visual servoing frameworks has been explored, with Visual Predictive Control (VPC)~\cite{10858477,li2021image}  emerging as a promising approach. By combining predictive modeling with image feedback, VPC enables proactive and flexible motion planning for the laparoscope. Meanwhile, Control Barrier Function (CBF)-based methods~\cite{wabersich2022predictive,2023Occlusion} have been introduced to explicitly encode FoV safety constraints, ensuring that the target remains visible and avoiding excessive FoV jitter that can impair the surgeon’s visual experience. However, both VPC and CBF approaches fundamentally depend on accurate system models~\cite{9838184}; In the presence of complex disturbances, their FoV safety guarantees may be significantly weakened, leaving unresolved robustness concerns. These disturbances are inherently heterogeneous, comprising deterministic, stochastic, periodic, and abrupt components, whose state-dependent characteristics make precise modeling intractable and simplistic norm-bounded assumptions overly conservative. In the broader robotics community, chance-constrained formulations have been extensively studied for path planning under uncertainty. A seminal work by Blackmore \cite{blackmore2011chance} introduced a probabilistic optimal planning framework that explicitly bounds the probability of constraint violation in stochastic environments. While this paradigm has proven effective, extending it to laparoscopic control requires fundamentally different considerations, as visibility safety must be enforced under task-specific FoV constraints and mixed disturbances that are absent in conventional planning settings.

To overcome the above limitations, this paper presents a robust autonomous control framework for robot-assisted laparoscope motion planning that ensures FoV safety under complex uncertainties. In this work, Gaussian Process Regression (GPR) is incorporated due to its nonparametric probabilistic structure and its ability to provide predictive uncertainty, enabling quantification of residual model errors, stochastic perturbations, and external disturbances~\cite{hewing2019cautious,10094232}. Based on this uncertainty characterization, a novel safety-aware trajectory optimization framework with probabilistic guarantees is formulated where an uncertainty-adaptive CBF condition via an uncertainty propagation, and chance constraints are formulated through probabilistic approximation. This uncertainty-aware formulation allows adaptive control effort allocation, suppressing unnecessary camera motion, and maintains robust FoV safety. The framework is validated through comparative simulations and hardware experiments on a commercial surgical robot platform (MicroPort MedBot$^\circledR$ Toumai$^\circledR$) performing sequential multi-target lymph node dissections. The main contributions are summarized as follows:


\begin{itemize}
\item [(1)]{Development of an integrated VPC framework (as shown in Fig.~\ref{framework}) for laparoscope-holding robots that ensures FoV safety under uncertainty. The framework is designed as a unified control architecture, where predictive trajectory planning and CBF-based safety guarantees are co-designed to achieve proactive and coordinated FoV management in both single- and multi-target surgical tasks, thereby maintaining continuous instrument visibility in procedures such as lymphatic dissection.}
\item [(2)]{Formulation of a principled uncertainty-adaptive control scheme grounded in Gaussian Process Regression (GPR). By leveraging GPR for hybrid (deterministic + stochastic) quantification of residual dynamics, stochastic disturbances, and external perturbations, the framework establishes uncertainty-propagated CBF conditions and chance-constrained safety formulations. This principled integration provides probabilistic safety guarantees, enables adaptive control effort allocation, and effectively suppresses FoV jitter.} 
\item [(3)]{Comprehensive validation through numerical simulations and hardware experiments on the MicroPort MedBot$^\circledR$ Toumai$^\circledR$ platform. Comparative studies against Classical VPC~\cite{9126170}, FullyTracking VPC~\cite{li2021image}, and VisionSafe VPC (FoV-constrained but without residual uncertainty handling) confirm that the proposed framework consistently achieves superior robustness and near-perfect visibility in dynamic instrument tracking and multi-target lymph node dissection tasks. The experimental video is available at: \url{https://youtu.be/ZhFDg9Ukt8Q}.}
\end{itemize}

The remainder of this paper is organized as follows: In Section~\ref{II}, problem statement is presented. In Section~\ref{III}, the VisionSafeEnhanced VPC method is elaborated. In Section~\ref{IV}, the results of numerical simulations and experiments are presented. In Section~\ref{V}, a conclusion is presented and potential avenues for future research are delineated.

\section{System Modeling} \label{II}
The proposed control framework is established on the kinematic model of the laparoscope-holding robotic arm and incorporates GPR for residual disturbance learning and uncertainty quantification.

Let $q\in \mathbb{R}^{6}$ denote the joint angles of the laparoscope-holding robotic arm, and let $\dot{x}=[v_{x},v_{y},v_{z},w_{x},w_{y},w_{z}]^{T}$ represent the linear and angular velocities of the Tool Center Point (TCP) in Cartesian coordinates,  corresponding to the end-effector velocity of the laparoscope (camera). The relationship joint and TCP velocities is given by the Jacobian matrix $J_r\in\mathbb{R}^{6\times6}$~\cite{2010Robotics}:
\begin{equation}\label{EQ1}
\dot{x}=J_r(q)\dot{q},
\end{equation}
\noindent where $J_r$ is determined by the geometric parameters and joint configuration of the robotic arm, and $\dot{q}$ denotes the joint velocities. 

A monocular camera mounted at the laparoscope tip observes surgical targets. A 3D spatial point $p=[x,y,z]^T$ in the camera coordinate projects onto the image plane as a 2D pixel coordinate $s=[u,v]^T$. The timederivative of the image features, $\dot{s}$, is related to the camera velocity $\dot{x}$ via the image interaction matrix $L_{s}\in\mathbb{R}^{2\times6}$~\cite{chaumette2006visual}:
\begin{equation}\label{EQ2}
\dot{s}=L_{s}\dot{x},
\end{equation}
with
\begin{equation}\label{EQ3}
L_s=
\begin{bmatrix}
\frac{-f_x}{d} & 0 & \frac{\hat{u}}{d} & \frac{\hat{u}\hat{v}}{f_y} & \frac{-\hat{u}^2}{f_x}-f_x & \frac{f_x\hat{v}}{f_y} \\
0 & \frac{-f_y}{d} & \frac{\hat{v}}{d} & \frac{\hat{v}^2}{f_y}+f_y & \frac{-\hat{u}\hat{v}}{f_x} & \frac{-f_y\hat{u}}{f_x}
\end{bmatrix},
\end{equation}
\begin{equation}\label{EQ31}
\hat{u}=u-u_0,\hat{v}=v-v_0,
\end{equation}

\noindent where $f_x$ and $f_y$ denoting the camera focal lengths, $[u_0,v_0]^T$ the principal point coordinates, and $d$ the target depth in the camera frame.

Combining Eq.~(\ref{EQ1}) and Eq.~(\ref{EQ2}) yields the mapping from joint velocities to image feature velocities:
\begin{equation}\label{EQ4}
\dot{s}=J_s\dot{q},
\end{equation}
\noindent where $J_{s}=L_{s}J_{r}$ is the image Jacobian. In practical laparoscopic surgical systems, this kinematic relationship is affected by multiple sources of uncertainty due to strong coupling between joint, task, and image spaces. Internal disturbances, including joint friction, cable-driven compliance, and elastic deformation of instruments~\cite{wei2024composite}, introduce nonlinear and time-varying residual dynamics that cannot be fully captured by the nominal model. Such factors degrade the accuracy of conventional visual servoing methods that rely on precise kinematic models.

To account for these effects, the system dynamics are expressed and discretized via the Euler method as:
\begin{equation}\label{EQ5}
s(k+1)=s(k)+\delta_{t}J_{s}\dot{q}(k)+B_{d}(d(s(k),\dot{q}(k))+w(k)),
\end{equation}
\noindent where \( s(k) \) represents the image feature position at time step $k$, \( \dot{q}(k) \) is the control input (desired joint velocity), \( \delta_t \) is the sampling period, and \(B_{d}\) is the disturbance distribution matrix. The model consists of a nominal term \(J_{s}\) and an additive residual term \(d(s(k),\dot{q}(k))\) representing unmodeled nonlinear dynamics within the subspace spanned by \(B_{d}\). The process noise
\( w(k) \sim \mathcal{N}(0, \sigma_\omega^2) \) is assumed i.i.d. and spatially uncorrelated.

GPR is employed as a nonparametric Bayesian  approach to model the residual disturbance term \(d(s(k),\dot{q}(k))\) from historical data, while simultaneously providing an explicit measure of predictive uncertainty~\cite{wei2023contact}. In the context of lymph node dissection, training data—including joint velocity, image feature positions, and tracking errors—are utilized to construct the GPR model. The predictive mean can be leveraged for real-time disturbance compensation,  while the predictive variance can serve as a quantitative measure of model confidence. Unlike parametric approaches, GPR does not require explicit structural assumptions about the residual dynamics and inherently provides both mean and variance estimates. These properties make GPR well-suited for enabling uncertainty-aware control, with the potential to enhance robustness and maintain continuous synchronization between the instruments and the FoV under dynamic surgical conditions.

\section{Control Framework Based on FoV Safety}\label{III}
This section presents a predictive control framework integrating CBFs with GPR to ensure FoV safety and enhance robustness in autonomous laparoscope motion. The CBF module imposes FoV safety constraints, preventing visibility loss during procedures such as lymph node dissection, while the GPR module models residual disturbances and quantifies predictive uncertainty, enabling cautious and accurate motion control.

\subsection{Constrained Predictive Control}\label{sec3.1}
The control objective is to keep the image features $s(k)$ within a safe FoV set $\mathcal{C}^{*}$, while tracking the target $s_d(k)$ and respecting joint limits. For the discrete-time system in Eq.~(\ref{EQ5}), the control input is defined as $u(k) \stackrel{\triangle}{=} \dot{q}(k)$, i.e., the joint velocity command of the laparoscope-holding robotic arm.
The VPC problem minimizes:
\begin{equation}\label{EQ11}
J = \sum_{i=1}^{N_p} (s_{k+i}-s_d)^T Q (s_{k+i}-s_d) 
  + \sum_{i=0}^{N_c} \dot{q}_{k+i}^T R \dot{q}_{k+i},
\end{equation}
subject to
\begin{align*}
q_{\min} \leq q_{k+i} \leq q_{\max}, \quad
\dot{q}_{\min} \leq \dot{q}_{k+i} \leq \dot{q}_{\max},
\end{align*}
where $Q$ and $R$ are positive definite weight matrices penalizing tracking error and control effort, respectively. $N_p$ and $N_c$ are the prediction and control horizons.

To guarantee visibility, let the FoV-safe set be
\begin{equation}\label{EQ13}
\mathcal{C}^{*}(k) = \{ s(k) \in \mathbb{R}^2 \mid h(s(k)) \ge 0 \},
\end{equation}
with
\begin{equation*}
    h(s(k)) = r^2 - \| s(k) - s_c \|^2,
\end{equation*}
where $s_c\!\in\!\mathbb{R}^2$ is the FoV center and $r\!>\!0$ is the nominal radius (parameterized by camera intrinsics). $h(s(k))$ measures the distance from the tip of the instrument to the FoV
center under ideal conditions. Forward invariance of $\mathcal{C}^{*}$ is enforced by the discrete-time CBF condition~\cite{ames2019control}
\begin{equation}\label{EQ114}
\mathcal{CBC}^{*}(k) \stackrel{\triangle}{=} h(s(k+1)) - (1-\alpha)\,h(s(k)) \;\ge\; 0,
\end{equation}
where $0 < \alpha \le 1$, and \(\mathcal{CBC}^{*}(k) = \mathcal{CBC}^{*}(s(k), \dot{q}(k))\), which guarantees that if $s(k)\!\in\!\mathcal{C}^{*}_{k}$ then $s(k+i)\!\in\!\mathcal{C}^{*}_{k}$ for all $i\!\ge\!0$. 

\subsection{Disturbance Learning and Uncertainty Quantification via GPR}
Residual dynamics $d(s(k),\dot{q}(k))$ are learned offline using GPR from experiments on the MicroPort Toumai platform (varied joint motions and simulated scenarios). With inputs $z_k=[s_k^{T},\dot{q}_k^{T}]^T$ and outputs computed as
\begin{equation}\label{EQy}
y_k \;=\; B_d^{\dagger}\big(s_{k+1} - s_k - \delta_t J_s \dot{q}_k\big),
\end{equation}
\noindent where \(B_d^\dagger\) is the pseudo-inverse of the disturbance mapping matrix,
GPR yields the predictive distribution
\begin{equation}
p((d(s_{k}, \dot{q}_{k}) + w_{k}) | \mathcal{D}, z_k) = \mathcal{N}(\mu(z_k), \Sigma_d(z_k)),
\end{equation}
where the training set \(\mathcal{D} = \{Z, Y\}\) consists of input-output pairs capturing the dynamics of the residual dynamics, and $\mu(z_k)$ estimates the residual disturbance and $\Sigma_d(z_k)$ quantifies predictive uncertainty (confidence), which are computed as follows
\begin{subequations}
\begin{align}
\mu(z_k) &= k(z_k, Z) \Lambda^{-1} Y, \\
\Sigma_d(z_k) &= k(z_k, z_k) - k(z_k, Z)\Lambda^{-1} k(Z, z_k),
\end{align}\label{EQgp}
\end{subequations}
\noindent where $\Lambda = K_{ZZ} + \sigma_n^2 I$, and $\sigma_n^2$ denotes the variance of the observation noise and $[K_{ZZ}]_{ij} = k(z_i, z_j)$, which denotes a squared-exponential kernel used through off-line training
\begin{equation}
k(z_i, z_j) = \sigma_f^2 \exp\left(-\frac{1}{2} (z_i - z_j)^T L^{-1} (z_i - z_j)\right),\label{kk}
\end{equation}
\noindent where $\sigma_f^2$ represents the signal variance of the squared exponential kernel, \(L = \text{diag}(l_1, \ldots, l_8)\) is the length-scale matrix corresponding to the eight-dimensional input space (two image features and six joint angles). 

Online, the one-step state predictor uses the mean compensation:
\begin{equation}\label{EQpred}
\bar{s}_{k+1} \;=\; \bar{s}_k + \delta_t J_s \dot{q}_k + B_d\,\mu(z_k),
\end{equation}
and the propagated image-space disturbance covariance is
\begin{equation}\label{EQcov}
\Sigma_{k+1} \:=\; \Sigma_{k}+B_d\,(\Sigma_d(z_k)+\sigma_\omega^2 I)\,B_d^T,
\end{equation}
\noindent where $\sigma_\omega^2$ is the process-noise variance and $I$ the identity of appropriate size. One may replace $I$ by a learned diagonal if componentwise variances are available.
\subsection{Uncertainty-Adaptive CBF and Chance Constraints}
To explicitly incorporate GPR confidence into safety, the FoV margin is tightened according to the predicted uncertainty. Define an uncertainty-adaptive safe function
\begin{equation}\label{EQhs}
h_\sigma\!\big(\bar{s}(k),z_k\big) \;=\; \big(r - \beta\,\sigma_s(z_k)\big)^2 - \|\bar{s}(k) - s_c\|^2,
\end{equation}
where $\beta\!\ge\!0$ scales uncertainty into a pixel margin and $\sigma_s(z_k)$ is an image-space standard deviation derived from $\Sigma_{k+1}$, as follows
\begin{equation}\label{EQsigmaproj}
\sigma_s(z_k) \;=\; \sqrt{\,\lambda_{\max}\!\big(\Sigma_{k+1}\big)\,}, 
\end{equation}
\noindent where $\lambda_{\max}(\cdot)$ denotes the maximum eigenvalue of a symmetric positive semi-definite matrix, corresponding to the largest principal variance direction in the image-space uncertainty covariance $\Sigma_{k+1}$. A sufficient (deterministic) CBF condition under online prediction uncertainty is then given by
\begin{equation}\label{EQcbc_sigma}
\mathcal{CBC}(k)\stackrel{\triangle}{=}h_\sigma\!\big(\bar{s}_{k+1},z_k\big) \;-\; (1-\alpha)\,h_\sigma\!\big(\bar{s}(k),z_k\big) \;\ge\; 0,
\end{equation}
by this way, the nominal next state $\bar{s}_{k+1}$ (mean prediction Eq.\eqref{EQpred}) must respect a \emph{shrunk} FoV radius $r-\beta\sigma_s$ that scales with the predicted uncertainty. This implements an uncertainty-aware safety margin: larger variance $\sigma_s$ yields a tighter feasible region.

In addition, chance constraints for state and input are imposed using the Gaussian approximation:
\begin{equation}\label{EQcc_state}
\Pr\big(\bar{s}_{k}\in\mathcal{C}_{k}\big) \;\ge\; p_s
\;\;\Longleftarrow\;\;
\mathcal{E}_s  \;+\; \kappa_s\sigma_s(z_k) \;\le\; r,
\end{equation}
\begin{equation}\label{EQcc_input}
\Pr\big(u_{k} \in \mathcal{U}_{k}\big) \;\ge\; p_u
\;\;\Longleftarrow\;\;
\mathcal{E}_u
+ \kappa_u\,\sigma_{u}(z_{k})\;\le\; r_u,
\end{equation}
where $\mathcal{E}_s = \|\bar{s}_{k}-s_c\|$, $\mathcal{E}_u =\| \bar{u}_{k} - u_c \|  $, $\kappa_s \stackrel{\triangle}{=}  \Phi^{-1}(p_s)$ is the standard normal quantile, and $\Phi^{-1}(\cdot)$ denotes the inverse cumulative distribution function. 
For the input constraint~\eqref{EQcc_input},  $\bar{u}_{k}$ is the nominal input prediction from the VPC, $u_c$ is the center of the admissible velocity set $\mathcal{U}_{k}$ (typically the midpoint of $\dot{q}_{\min}$ and $\dot{q}_{\max}$), and $r_u$ is its admissible radius. $\sigma_{u}(z_{k})=\sqrt{\lambda_{\max}(\Sigma^u_{k})}$, where
$\Sigma^u_{k}$ is the control covariance, determined through an offline calibration process. $\kappa_u\stackrel{\triangle}{=}\Phi^{-1}(p_u)$ specifies the confidence level for input feasibility. These spectral-norm bounds yield Second Order Cone (SOC) tightenings that are convex and practically conservative, ensuring robustness against execution noise and modeling errors in the input.
\subsection{Integrated Optimization Problem}
Combining the above elements, the uncertainty-aware FoV safety optimization problem is formulated as
\begin{equation}\label{EQfinal}
\begin{aligned}
\min_{\{u_{k}\}_{i=0}^{N_c}} \ & 
\sum_{i=1}^{N_p} (\bar{s}_{k+i}-s_d)^T Q (\bar{s}_{k+i}-s_d) 
+\sum_{i=0}^{N_c} u_{k}^T R u_{k} \\[2pt]
\text{s.t. } \ & q_{\min} \le q_{k+i} \le q_{\max},\;\;\\[2pt]
& \text{Chance constraints } \eqref{EQcc_state} \text{--} \eqref{EQcc_input}, \\[2pt]
& \text{Uncertainty-adaptive CBF condition } \eqref{EQcbc_sigma}, \\[2pt]
& \bar{s}_{k+1} \;=\; \bar{s}_k + \delta_t J_s \bar{u}_k + B_d\,\mu(\bar{z}_k), \\
& \Sigma_{k+1} \;=\; \Sigma_{k}+B_d\,(\Sigma_d(\bar{z}_k)+\sigma_\omega^2 I)\,B_d^T,
\end{aligned}
\end{equation}
where $s_0=\bar{s}(k)$, and $\mu(\bar{z}_k),\Sigma_d(\bar{z}_k)$ are provided by GPR as in \eqref{EQgp}. $\bar{z}_k$ is the mean value of $z_{k}$ and is defined as $\bar{z}_k=[\bar{s}_{k}^{T},\bar{u}_{k}^{T}]^{T}$. The CBF condition is given in \eqref{EQhs}, with $\beta$ tuning the conservatism of the safety margin and $\alpha$ setting the CBF contraction rate~\cite{ames2019control}. The probability thresholds $p_s$ and $p_u$ in \eqref{EQcc_state}--\eqref{EQcc_input} specify the required confidence levels for state and input feasibility. This formulation fuses mean-based disturbance compensation with variance-aware safety tightening, thereby improving FoV retention and robustness under dynamic surgical conditions.
 
\section{Experimental Validation}\label{IV}
The proposed framework was validated through numerical simulations and hardware experiments. Simulations evaluated FoV safety and tracking precision under controlled conditions, while experiments on a commercial laparoscopic platform assessed real-world applicability.
\subsection{Numerical Simulations}
The simulated environment modeled the laparoscope’s distal FoV as a circular region. An active curved scissors executed a multi‑target sequence (L1 $\to$ L2 $\to$ L3 $\to$ L1) and was required to remain visible. Two aspects were examined: (i) \emph{FoV safety}, validating the CBF-based visibility constraint (VisionSafe VPC) as a core component; and (ii) \emph{robustness under disturbance}, evaluated by VisionSafeEnhanced VPC (VisionSafe VPC with uncertainty-adaptive CBF and cautious disturbance compensation).

To emulate practical uncertainties, a $10\%$ perturbation was injected into DH parameters, and additional process/measurement noises were modeled as zero-mean Gaussian with standard deviation $0.001\,\mathrm{m}$. The scissors were teleoperated along the multi-target sequence, while the laparoscope autonomously adjusted its pose to maintain visibility.

Three control schemes were evaluated: \emph{Classical VPC} (no visibility constraints), \emph{VisionSafe VPC} (CBF-based FoV safety), and \emph{VisionSafeEnhanced VPC} (cautious CBF + GPR-based disturbance learning). Metrics included: FoV Satisfaction Rate (FoVSR), representing the percentage of time the scissors remain within the FoV; the Mean Absolute Error (MAE) and Root Mean Square Error (RMSE), which quantify robustness against disturbances by measuring the deviation between the ground-truth trajectory (generated using the nominal disturbance-free model) and the trajectory obtained under disturbed conditions; and the Average Violation Severity Score (AVSS--average normalized deviation outside the FoV boundary),
\begin{equation}
\epsilon_i = \max\!\left(0, \frac{\|s_i - s_c\|}{r} - 1\right),\quad
E = \frac{1}{N} \sum_{i=1}^N \epsilon_i,
\end{equation}
where $s_i$ is the instrument image position, $s_c$ the FoV center, $\epsilon_i$ denotes the violation severity at time $i$, and $E$ its average over \( N \) time steps. In this case, a higher value corresponds to a greater deviation from the FoV center, reflecting reduced safety.

\begin{table}[!b]
\caption{Tracking Robustness Results}
\centering
\resizebox{\columnwidth}{!}{ 
\begin{tabular}{ccccc}
\toprule  
\textbf{Method} & \textbf{MAE} & \textbf{RMSE} & \textbf{FoVSR}  & \textbf{AVSS}  \\ 
\midrule
Classical VPC & -- & -- & 61.73\%  & 0.3174 \\ 
VisionSafe VPC & 0.0993 & 0.1065 & 98.97\% &  0.0020 \\ 
VisionSafeEnhanced VPC & 0.0224 & 0.0251 & 99.98\%  & 3.5E-5 \\ 
\bottomrule
\end{tabular}
}
\label{table2}
\end{table}

As summarized in Table~\ref{table2}, Classical VPC loses track early (FoVSR $=61.73\%$) because the laparoscope remains fixed and the scissors exit the FoV; MAE/RMSE are omitted since tracking is not sustained. VisionSafe VPC maintains high visibility (FoVSR $=98.97\%$) but exhibits noticeable deviations (MAE $=0.0993$, RMSE $=0.1065$) and higher AVSS ($2.0\times 10^{-3}$) under disturbances such as joint friction and elastic deformation. VisionSafeEnhanced VPC achieves the best performance, with FoVSR $=99.98\%$, the lowest errors (MAE $=0.0224$, RMSE $=0.0251$), representing a $77\%$ RMSE reduction over VisionSafe VPC, and minimal AVSS ($3.5\times 10^{-5}$).

\begin{figure}[!t]
\centering 
\includegraphics[width=1\linewidth]{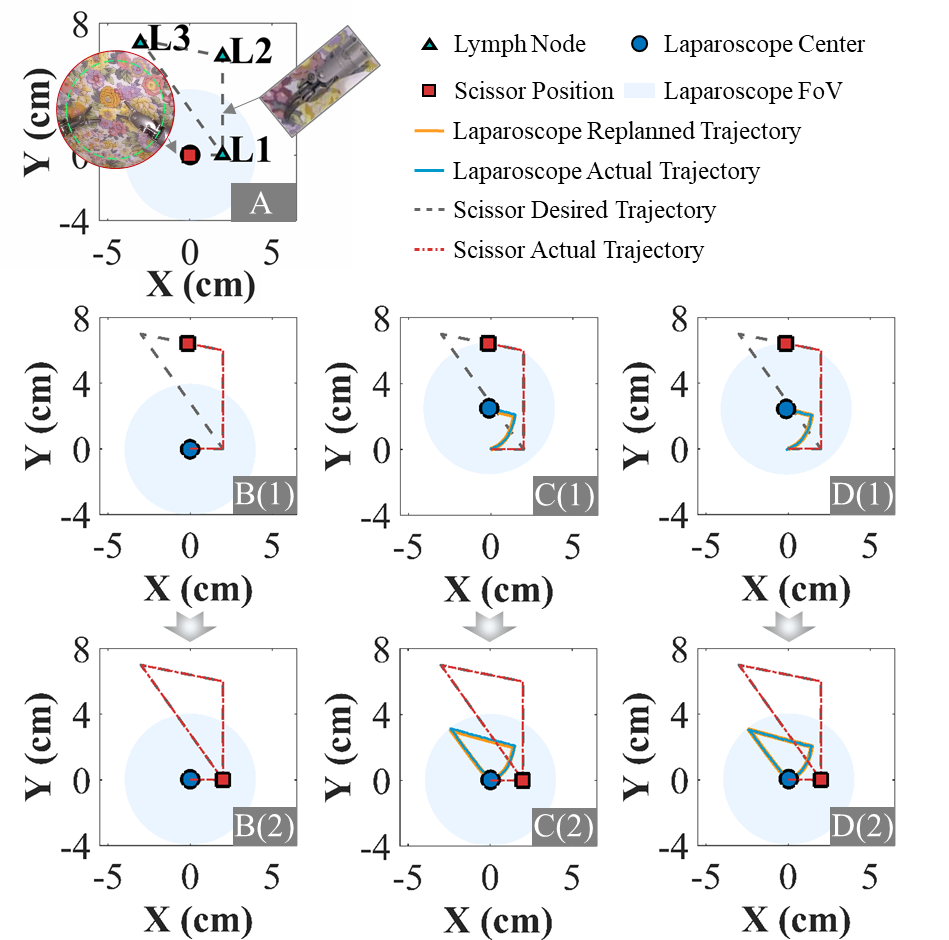}
\caption{Comparison of Tracking Accuracy for Active curved scissors and Static Laparoscope Motion: (A) Initial state. (B) Classical VPC: the laparoscope remains fixed and the scissors leave the FoV. (C) VisionSafe VPC: scissors stay within the FoV but the executed trajectory deviates from the replanned one under disturbance. (D) VisionSafeEnhanced VPC: high tracking accuracy with executed and replanned trajectories closely overlapping.}
\label{fig5}
\end{figure}

Fig.~\ref{fig5} visualizes these results. Panel A shows the initial configuration. In Classical VPC (B(1)–B(2)), the laparoscope remains stationary, causing the scissors to leave the FoV. VisionSafe VPC (C(1)–C(2)) replans the laparoscope path when the scissors approach the visibility boundary, keeping them nominally within the FoV; however, disturbance effects cause the executed trajectory (red dash-dotted) to deviate from the replanned one (yellow), occasionally breaching the FoV. VisionSafeEnhanced VPC (D(1)–D(2)) leverages GPR to learn and compensate residual disturbances online, resulting in executed trajectories (red) closely overlapping replanned ones (yellow) and maintaining continuous visibility throughout the multi-target sequence. These qualitative observations align with the quantitative gains in Table~\ref{table2}, confirming that explicit FoV constraints prevent visibility loss, while learned disturbance compensation is essential for high-precision, robust tracking in dynamic surgical scenarios.

\begin{figure}[!t]
	\centering
\includegraphics[width=1\linewidth]{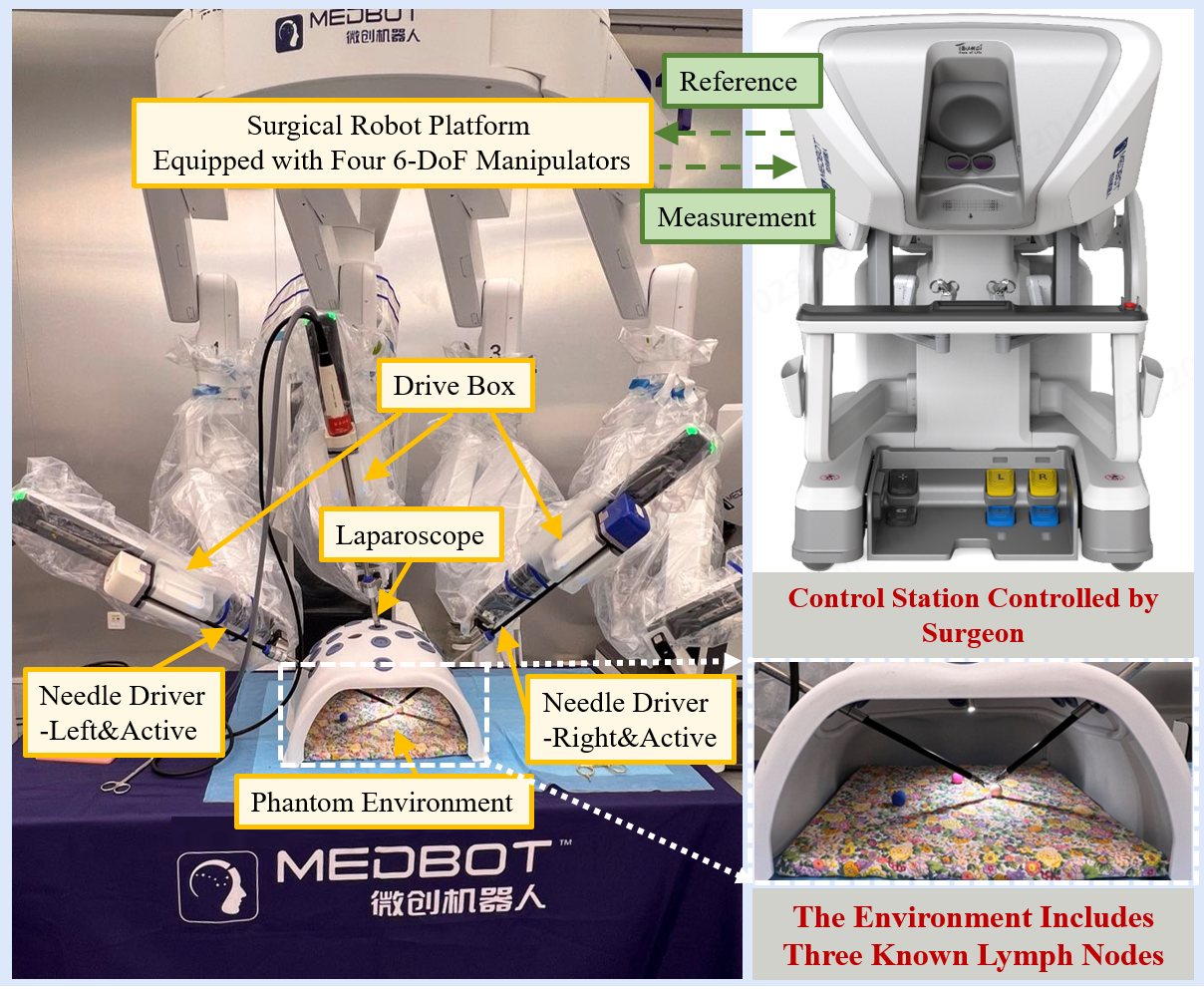}
\caption{The experimental setup.}
\label{fig2}
\end{figure}
\subsection{Experiments in multi-target Surgical Scenarios}
\subsubsection{Experimental setup}
The proposed VisionSafeEnhanced VPC framework was experimentally validated on the MicroPort Toumai surgical robot platform (Fig.~\ref{fig2}), which integrates a stereoscopic laparoscope and four robotic manipulators, representative of standard MIS configurations. The experimental task simulated a multi-target lymph node dissection, requiring precise coordination between instrument motion and laparoscopic view control. 
\begin{table}[!t]
\caption{Controller Parameters}
\label{table5}
\centering
\begin{tabular}{lll}
\hline
\textbf{Parameter Name} & \textbf{Symbol} & \textbf{Value} \\ \hline
Prediction Horizon & \( N_p \) & 10 \\ \hline
Control Horizon & \( N_c \) & 5 \\ \hline
Image Error Weight Matrices & \( Q \) & $I_2$ \\ \hline
Control Input Weight Matrices & \( R \) & $0.1\cdot I_6$ \\ \hline
CBF Parameter & \( \alpha \) & 0.5 \\ \hline
GPR Length-Scale Matrices & \( L \) & $0.5\cdot I_6$ \\ \hline
GPR Signal Variance & \( \sigma_f^2 \) & 0.1 \\ \hline
\end{tabular}
\end{table}

A phantom model was prepared using a surgical drape with three sutured spheres (L1–L3) to emulate lymph nodes. Two needle drivers sequentially visited these targets from an initial position L0. To maintain situational awareness and surgical safety, a dynamically defined virtual target point $p_v$ was kept within the laparoscopic FoV, ensuring both the active instrument and the upcoming target were centered. The virtual point was calculated as
\begin{equation}
p_v = w_{n1} p_{n1} + w_{n2} p_{n2} + w_l p_l,
\end{equation}
where $p_{n1}$ and $p_{n2}$ are the instrument tips, $p_l$ is the current target, and weights $(w_{n1}, w_{n2}, w_l)$ are adaptively updated based on proximity to the target~\cite{gruijthuijsen2022robotic}. Controller parameters, tuned via simulation, are summarized in Table~\ref{table5}, including prediction horizon $N_p$, control horizon $N_c$, image error weight $Q$, control effort weight $R$, CBF parameter $\alpha$, and GPR hyperparameters (length-scale $L$, signal variance $\sigma_f^2$). For the GPR disturbance model, offline initialization used 50 samples, followed by online updates with a sliding window of the 30 most recent data points.
\begin{table}[!hbt]
\caption{Comparative performance matrices}
\centering
\label{tab:results}
\resizebox{\columnwidth}{!}{ 
\begin{tabular}{ccccccc}
\toprule  
\textbf{Method} & \textbf{MAE} & \textbf{RMSE} & \textbf{FoVSR} & \textbf{AVSS}  & \textbf{MAC} \\ 
\midrule
FullyTracking VPC & 0.1565 & 0.1878 & 100.00\% & -- & 68.3044 \\ 
VisionSafeEnhanced VPC & 0.0632 & 0.0758 & 99.98\%  & 0.0067 & 3.2886 \\ 
\bottomrule
\end{tabular}
}
\end{table}
\subsubsection{Experimental results}
\begin{figure}[!t]  \centering\includegraphics[width=1.0\linewidth]{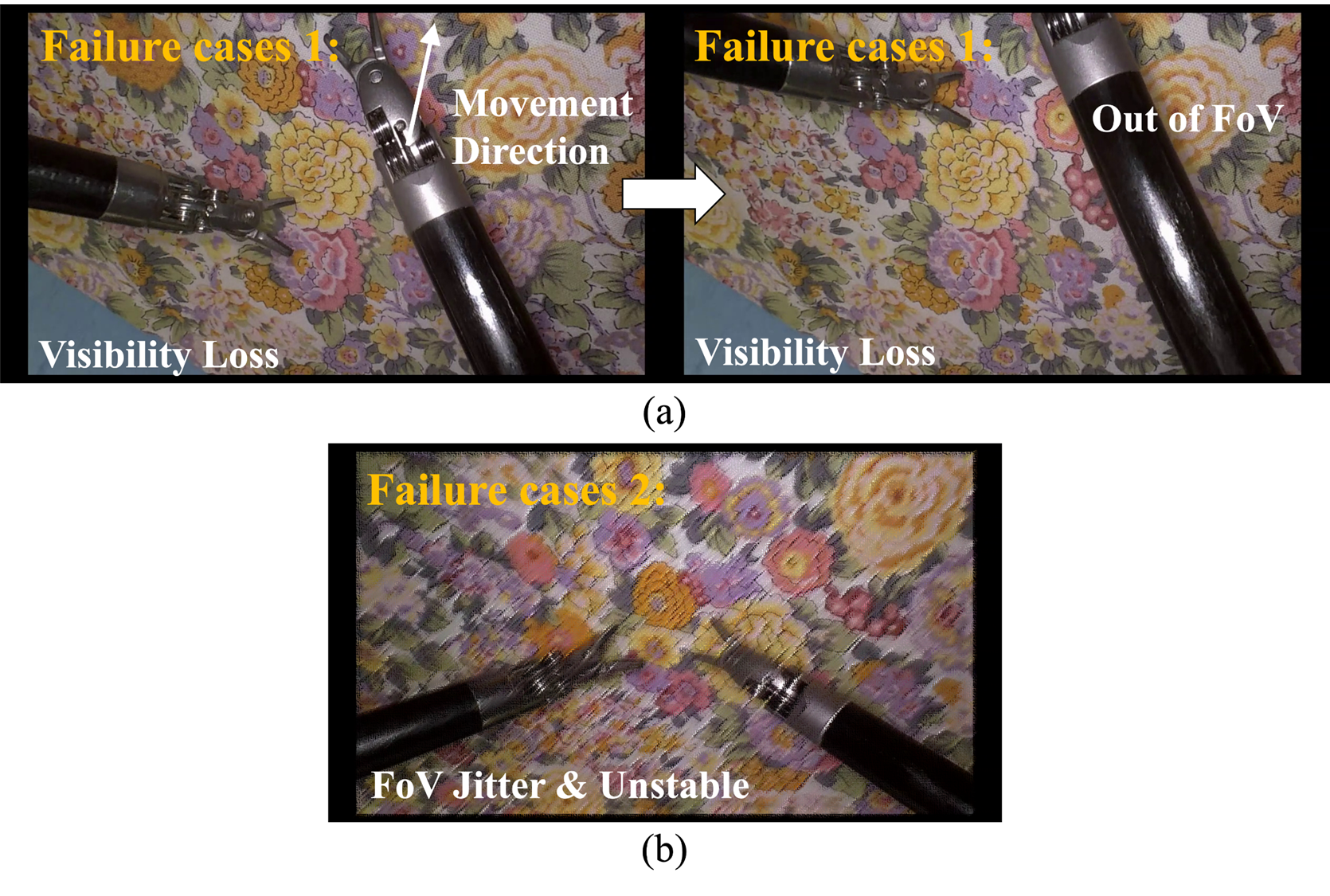}
\caption{Failure modes of two baseline methods. (a) The classical VPC fails as the instrument moves out of the FoV. (b) The FullyTracking VPC creates significant FoV jitter and an unstable view.}
    \label{fig:fig8}
\end{figure}

The first evaluation considered the L0$\to$L1 segment under three methods: \emph{Classical VPC} (no FoV constraints), \emph{FullyTracking VPC} (continuous IBVS-based centering~\cite{bista2016appearance}), and the proposed \emph{VisionSafeEnhanced VPC}. As shown in Fig.~\ref{fig:fig8}-a, Classical VPC lost visibility as instruments exited the FoV and was excluded from quantitative comparison. FullyTracking VPC (Fig.~\ref{fig:fig8}-b) maintained visibility but produced severe FoV jitter and excessive Motion Activity Cost (MAC), potentially inducing operator discomfort.  The MAC is defined as
\begin{equation}
E_k = \sum_{\tau=0}^k \dot{q}_{\tau}^\mathrm{T} M(q_\mathrm{ref}) \dot{q}_{\tau},
\end{equation}
where $M(q_\mathrm{ref})$ is the mass matrix at a reference configuration $q_\mathrm{ref}$, quantifying total motion as the time-integrated pseudo-kinetic energy. 
\begin{figure*}[!htb]
\centering 
\includegraphics[width=1\linewidth]{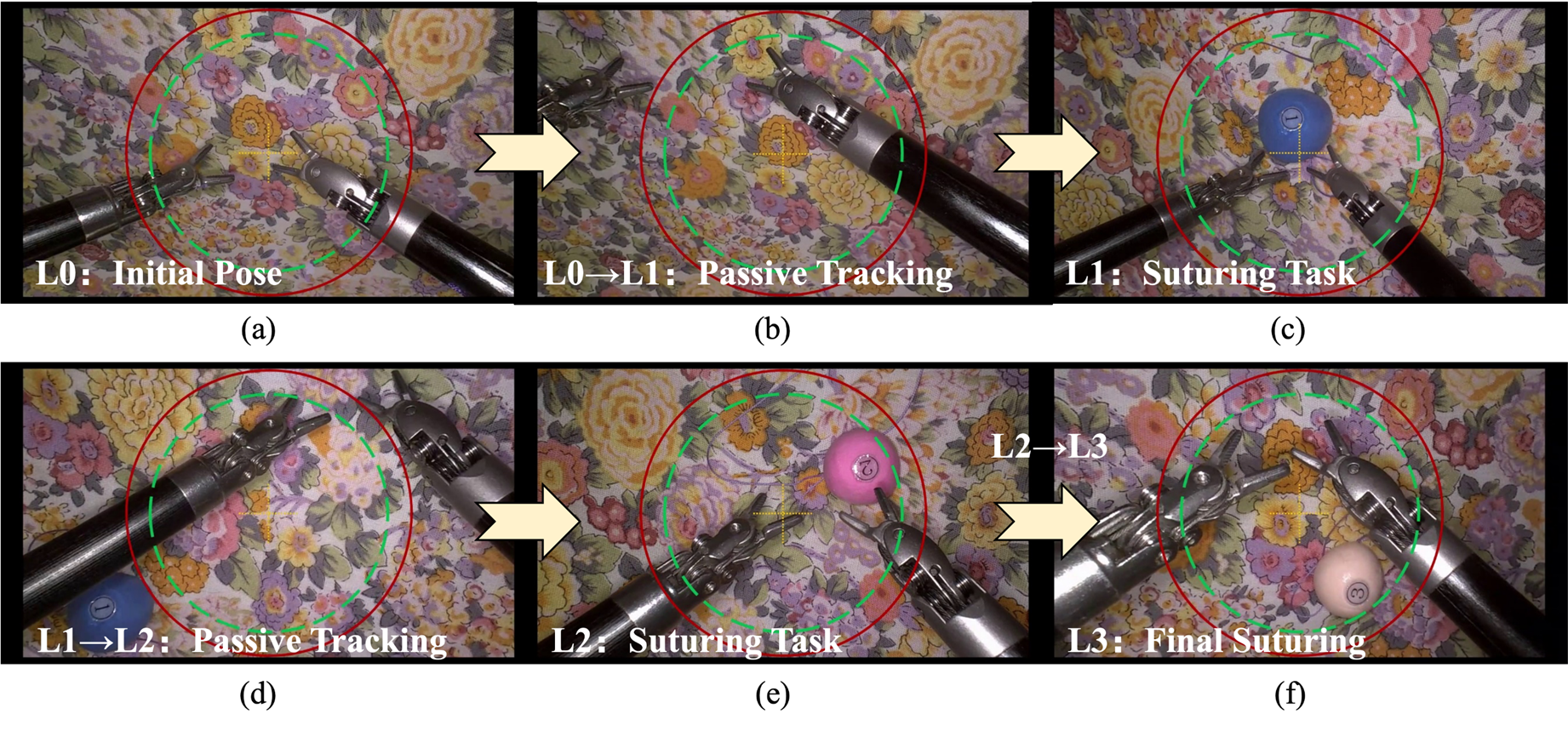}
\caption{System validation on a sequential multi-target lymph node dissection task. Starting from the initial pose (a), the controller alternates between passively tracking the instruments during transit to new targets (b, d) and maintaining a stationary view during suturing tasks at each location (c, e, f). Throughout the procedure, the target is kept within the safety constraint (green circle). The red circle is the FoV.}
\label{fig7}
\end{figure*}

\begin{figure}[!hbt]
\centering
\includegraphics[width=1.0\linewidth]{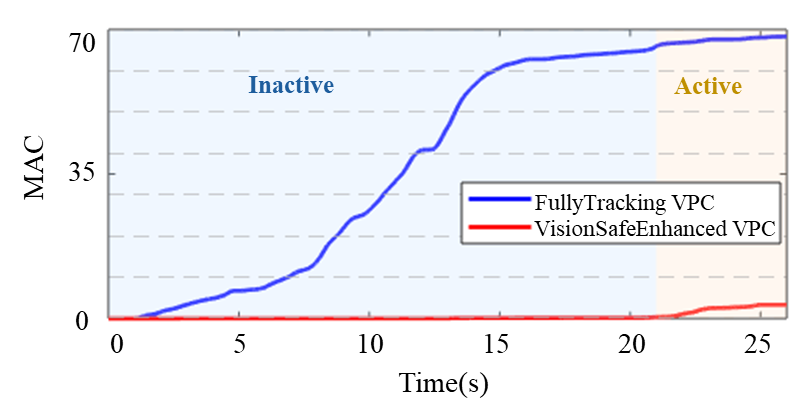}
\caption{MAC comparison for L0 $\to$ L1 segment. The shaded regions denote the two operational states of the proposed controller: the \emph{Inactive} state (light blue), where it remains passive, and the \emph{Active} state (light orange), where it becomes engaged.}
\label{fig:fig110}
\end{figure}

Quantitative results for this segment (as shown in Table~\ref{tab:results}) include: MAE, RMSE, FoVSR, AVSS, and MAC. As shown in Fig.~\ref{fig:fig110}, it is illustrated that FullyTracking VPC exhibited a steep, continuous MAC rise, reflecting persistent tracking effort, whereas VisionSafeEnhanced VPC maintained near-zero MAC for most of the task (inactive state, $t<21$\,s) and increased modestly only when approaching the FoV boundary (active state). Both methods achieved FoVSR $>99.9\%$, but VisionSafeEnhanced VPC reduced MAE and RMSE by $\approx59.6\%$ and lowered cumulative MAC by over 20 times, confirming its superior tracking accuracy and efficiency.

Building on the single-segment results, the full L0$\to$L1$\to$L2$\to$L3 multi-target sequence was executed to assess robustness in a clinically relevant, dynamic setting. Throughout the task, the controller alternated between passive observation during instrument transit and active laparoscope adjustment near the FoV boundary, as illustrated in Fig.~\ref{fig7}. The virtual target point $p_v$ remained consistently within the predefined safety margin (green circle), enabling uninterrupted execution of all target resections.

\begin{table}[!hb]
\caption{Quantitative evaluation for the full sequence}
\centering
\label{tab:results2}
\resizebox{\columnwidth}{!}{ 
\begin{tabular}{ccccccc}
\toprule  
\textbf{Method} & \textbf{MAE} & \textbf{RMSE} & \textbf{FoVSR} & \textbf{AVSS}  & \textbf{MAC} \\ 
\midrule
VisionSafeEnhanced VPC & 0.0658 & 0.0763 & 99.92\%  & 0.0089 & 14.798 \\ 
\bottomrule
\end{tabular}}
\end{table}

Quantitative results for the complete sequence (Table~\ref{tab:results2}) show a FoVSR of $99.92\%$, minimal AVSS (0.0089), and low tracking errors (MAE = 0.0658, RMSE = 0.0763), achieved under modeled disturbances including joint friction and instrument elasticity. These results demonstrate that the VisionSafeEnhanced VPC, leveraging CBF-enforced FoV safety and GPR-based disturbance learning, achieves high tracking precision, robust visibility maintenance, and substantially reduced unnecessary camera motion, thereby enhancing both surgical safety and operational efficiency in complex MIS scenarios.

\section{Conclusion} \label{V}
This paper presented VisionSafeEnhanced VPC, a robust control framework for autonomous laparoscope control in robotic MIS under visibility constraints and model uncertainty. The framework integrates VPC for trajectory optimization, CBF to enforce FoV safety, and GPR to learn and compensate for system disturbances. It is designed to prevent FoV loss under dynamic conditions, reduce excessive responsiveness for enhanced surgeon comfort, and improve control accuracy by compensating for system uncertainties.

The framework's efficacy was systematically validated through numerical simulations and physical experiments on the MicroPort Toumai surgical robot platform. Comparative studies demonstrated that the CBF-based constraints are critical for safety, achieving a $100\%$ FoV satisfaction rate where classical methods fail. On a multi-target lymph node dissection task, the complete framework achieved $99.92\%$ target visibility. Furthermore, the GPR-based compensation was shown to significantly enhance robustness against parametric uncertainties, reducing tracking error by over $77\%$ compared to the CBF-only approach.

While the proposed framework excels in dynamic tracking and FoV assurance, it can directly contribute to improved procedural safety, reduced surgeon fatigue, and higher precision in automated tasks. Future research will extend this work in several directions. First, we will investigate tighter integration with surgeon-in-the-loop systems, developing intelligent interfaces that adapt to operator intent for more intuitive human-robot collaboration. Furthermore, the current safety logic will be evolved into more flexible constraints that can be modulated based on the surgical context.

\ifCLASSOPTIONcaptionsoff
  \newpage
\fi

\footnotesize
\bibliographystyle{IEEEtran}
\bibliography{RAL_23_3298}

\end{document}